\documentclass[runningheads]{llncs}
\usepackage{amsmath}
\usepackage{amsfonts}
\usepackage{centernot}
\usepackage{booktabs}
\usepackage{multirow}
\usepackage[bookmarks]{hyperref}
\newcommand\given[1][]{\:#1\vert\:}
\usepackage{graphicx}
\graphicspath{{images/}}
\begin{document}
\title{A cross-center smoothness prior for\\ variational Bayesian brain tissue segmentation}
\titlerunning{A cross-center prior for variational Bayesian tissue segmentation}
%
\author{Wouter M. Kouw\thanks{Supported by a contribution from the Niels Stensen Fellowship.}\inst{1} \and
Silas N. \O rting\inst{1} \and
Jens Petersen\inst{1}\\  \and
Kim S. Pedersen\inst{1} \and
Marleen de Bruijne\inst{1,2}
}
\authorrunning{W.M. Kouw et al.}
%
\institute{University of Copenhagen, Universitetsparken 1
DK-2100 Copenhagen \O, Denmark \and
Erasmus Medical Center, Dr. Molewaterplein 50, 3015 GE Rotterdam, Netherlands
}
\maketitle              
\begin{abstract}
Suppose one is faced with the challenge of tissue segmentation in MR images, without annotators at their center to provide labeled training data. One option is to go to another medical center for a trained classifier. Sadly, tissue classifiers do not generalize well across centers due to voxel intensity shifts caused by center-specific acquisition protocols. However, certain aspects of segmentations, such as spatial smoothness, remain relatively consistent and can be learned separately. Here we present a smoothness prior that is fit to segmentations produced at another medical center. This informative prior is presented to an unsupervised Bayesian model. The model clusters the voxel intensities, such that it produces segmentations that are similarly smooth to those of the other medical center. In addition, the unsupervised Bayesian model is extended to a semi-supervised variant, which needs no visual interpretation of clusters into tissues.
\end{abstract}
\section{Introduction}
Many modern automatic brain tissue segmentation methods are based on machine learning models. One of the limitations of these models is that they generalize poorly beyond the domain of the data they are trained on. In medical imaging, an example of a domain is the medical center itself. Data collected at different centers varies due to experimental, acquisition and annotation protocols. Most notably, the voxel intensity distributions of MR images are different, which means the mapping from scans to segmentations differs between centers. As a result, tissue classification models trained on examples from one center tend to perform poorly on data from another center \cite{van2015transfer}.

But not all factors of variation are inconsistent across centers. Although the segmentations are different for each patient, certain aspects remain consistent. One such aspect is the spatial \emph{smoothness} of each tissue. Radiologists and MR imaging experts know how smooth a segmentation is supposed to look like, and use this knowledge when segmenting a new scan. Essentially, we would like to give the tissue classifier that information as well. Our goal is to learn from segmentations at other medical centers and incorporate that knowledge into a Bayesian model for tissue segmentation.

\subsection{Related Work}
In transfer learning and domain adaptation, a model learns from a \emph{source} domain and aims to generalize to a differently distributed \emph{target} domain \cite{pan2010survey,kouw2019review}. In Bayesian transfer learning, the source domain can be interpreted as prior knowledge for the target task \cite{raina2006constructing,finkel2009hierarchical}. For instance, in natural language processing, a document classification task can be performed using a Bayesian linear classifier trained on a bag-of-word encoding of the document \cite{raina2006constructing}. Instead of imposing a weakly informative prior on how important each word of the dictionary is for the document classification task, one could fit the prior on data from Wikipedia. That produces a stronger, more informative prior over how important each word is. To our knowledge, no Bayesian transfer learning models have been proposed for medical imaging tasks. Our interest is to study what forms of prior knowledge can be obtained from large open access labeled data sets, and how that knowledge can be exploited for a specific task.

Hidden Markov Random Field (MRF) models are a form of Bayesian models for image segmentation. They pose a hidden state for each voxel that accounts for some intrinsic latent structure of the image \cite{wang2013markov}. For tissue segmentation, the latent state is assumed to be the tissue of the voxel, while the observed voxel intensity value is a sample from a probabilistic observation model. The observation model specifies the causal relations between the latent image and the observed image \cite{zhang2001segmentation,ashburner2005unified}. Such assumptions are not unreasonable for the case of MR imaging, where T1 relaxation times depend on the tissue of the voxel. 

Inference in hidden MRF's is often done through Monte Carlo sampling \cite{winkler2012image}. However, sampling remains a computationally expensive procedure. An alternative is to use variational inference, where the joint distribution of an intractable Bayesian model is approximated \cite{bishop2006pattern,blei2017variational}. Variational inference is often much faster than sampling, depending on the form of the approximating distribution. We will employ variational Bayes to infer the underlying tissues of an observed MRI scan.

\subsection{Outline}
In Section \ref{sec:method}, we will discuss a Bayesian model for tissue segmentation along with a hidden Markov Random Field prior. The variational approximation and the general inference procedure is presented in Section \ref{sec:model}. Section \ref{sec:cross} covers how to fit the MRF prior to segmentations produced at other medical centers. We perform a series of experiments in Section \ref{sec:experiments} where we pair up data sets from different medical centers. Model extensions and limitations are discussed in Section \ref{sec:discuss} and we draw conclusions in Section \ref{sec:conclusion}.

\section{Method} \label{sec:method}
Let $X \in [0, 1]^{H \times W \times D}$ be an MR image and $Y \in \{0, 1\}^{H \times W \times K}$ be its segmentation. $H$ and $W$ are the height of the width of the image, respectively, with $N = H \cdot W$ as the total number of voxels. $D$ refers to the number of channels of the image, which could be stacked filter response maps or additional modalities. In this paper, we consider only the MR image (i.e. D=1), but the update equations in Section \ref{sec:vi} are general. $K$ corresponds to the number of tissues in the segmentation, also referred to as classes. Observed voxels are marked as $x$. Voxel labels are marked as $y$ and consist of $\{0,1\}$-valued vectors with $1$ on the $k$-th index if that voxel belongs to class $k$ (a.k.a. \emph{one-hot} vectors).

\subsection{Bayesian model} \label{sec:model}
We assume a causal model $Y \rightarrow X$, such that the tissue causes the voxel intensity value. The measurement instrument, i.e. the MRI scanner, maps tissues to voxel intensities $f$, but imposes noise on the observation: $x = f(y) + \epsilon$. The mapping $f$ between $Y$ and $X$ is assumed to vary across experimental and acquisition protocols. We model the likelihood function of observing $X$ from $Y$ with a Gaussian mixture model, with one component for each tissue:
\begin{align}
	p(X \given Y; \ \pi, \mu, \Lambda) = \ \prod_{i=1}^{N} \prod_{k=1}^{K} \big[ \pi_{k} \ \mathcal{N}(x_i \given \mu_{k}, \Lambda_{k}^{-1}) \big]^{y_{ik}} \, . \label{eq:likelihood}
\end{align}
The parameter $\pi_k$ is the proportion coefficient, $\mu_k$ is the mean intensity and $\Lambda_k$ is the precision of the $k$-th tissue. Note that this likelihood assumes that voxels are independent of each other, which is not valid in MR images. We model spatial relationships in Section \ref{sec:hpotts} which introduces dependencies between pixels. 

We select a Dirichlet distribution as the prior for the tissue proportions and a Normal-Wishart as the prior for the mean and precision parameters:
\begin{align}
    \pi_k \sim \ {\cal D}( \alpha_{0k}) \, , \quad \quad
    \mu_k \sim \ {\cal N}( \upsilon_{0k}, \ (\gamma_{0k} \Lambda_k)^{-1}) \, , \quad \quad
    \Lambda_k \sim \ {\cal W}( \nu_{0k}, \Delta_{0k}) \, . \nonumber
\end{align}
The $\alpha_{0}$ are called the Dirichlet distribution's concentration parameters, $\upsilon_{0}$ the hypermeans, $\gamma_{0}$ are precision-scaling hyperparameters, $\nu_{0}$ are the degrees of freedom of the Wishart distribution and $\Delta_{0}$ are the hyperprecisions. These priors are conjugate to the Gaussian likelihood.

\subsection{Hidden Potts - Markov Random Field} \label{sec:hpotts}
Spatial properties of images can be described using Markov Random Fields. In general, MRF's describe interactions between nodes in a graph by defining a probability distribution -- to be precise, a Gibbs distribution -- over configurations of states at the nodes \cite{wang2013markov}. The Markov property allows us to model this distribution in terms of local interactions. The Ising model is a classical MRF model, which describes the pairwise interactions between a binary-valued image pixel and its direct neighbours (i.e. up, down, left, right). The Potts model is its multivariate extension, using $K$ states. 

We use the Potts model to capture how often a voxel's label is equal to the labels of its neighbours. In other words, how \emph{smooth} the segmentation is. The model incorporates a set of parameters, $\beta = (\beta_1, \dots, \beta_K)$, that explicitly describes each tissue's smoothness. By fitting a hidden Potts model to a series of segmentations, it can act as an informative prior in the Bayesian model -- a point we discuss in more detail in Section \ref{sec:cross}. 

Hidden Potts models are usually defined for whole images. However, that induces a partition function with a discrete sum over $K^N$ states, which is computationally intractable. Instead, we consider a local variant, where voxels depend only on their direct neighbours $\delta_{i}$ \cite{mcgrory2009variational,liu2013image}:
\begin{align}
    p(Y \given \beta) \ = \ \prod_{i=1}^{N} \ p(y_{i} \given y_{\delta_{i}}, \beta) \, . \nonumber
\end{align}
Voxels in the center of the image have four neighbours (i.e. up, down, left, right), while edge and corner voxels have three and two neighbours, respectively. Taking its logarithm, the Potts model has the following form:
\begin{align}
    \log p(y_{i} \given y_{\delta_{i}}, \beta) \ =&  \ \sum_{k=1}^{K} \beta_k y_{ik} \sum_{j \in \delta_{i}} y_{jk} - \log \sum_{\{y'\}} \exp \big(\sum_{k=1}^{K} \beta_k \ y'_{k} \sum_{j \in \delta_{i}}  y_{jk} \big) \label{eq:potts1} \\
    =& \ \sum_{k=1}^{K} \beta_k y_{ik} \sum_{j \in \delta_{i}}  y_{jk} - \log \sum_{k=1}^{K} \exp \big( \beta_k \sum_{j \in \delta_{i}} y_{jk} \big) \, . \label{eq:potts}
\end{align}
The sum with the subscript $\{y'\}$ in (\ref{eq:potts1}) denotes summing over all possible states of $y$ (i.e. $[1, 0, \dots 0], \ [0, 1, \dots, 0], \ \dots, \ [0,0, \dots, 1]$). Since $y$ is a one-hot vector, it multiplies the terms in the sum that involve the $k$-th tissue with $1$ and multiplies the other terms with $0$. All but one term drop out, which means the sum over $\{y'\}$ can be simplified to a sum over classes, as in (\ref{eq:potts}).

\subsection{Variational approximation} \label{sec:vi}
The hidden Potts-MRF describes spatial relationships in the segmentation and acts as a prior on the Gaussian mixture model. Including the hidden Potts model, the joint distribution of the full model becomes:
\begin{align}
    p(X, Y,&\ \pi, \mu, \Sigma \given \beta) \nonumber \\
    & = \ p(X \given Y, \pi, \mu, \Sigma) \ p(Y \given \beta) \
    p(\pi \given \alpha) \ p(\mu \given \upsilon, (\gamma \Lambda)^{-1}) \ p(\Lambda \given \nu, \Delta) \, . \label{eq:full}
\end{align}
In the following, the likelihood parameters are summarized as $\theta = (\pi, \mu, \Lambda)$. With the inclusion of the hidden Potts-MRF, the posteriors cannot be derived analytically. We perform a variational approximation of the joint distribution using a distribution over the segmentation and the likelihood parameters, $q(Y, \theta \given \beta)$ \cite{bishop2006pattern}. This approximation relates to the marginal log-likelihood as follows:
\begin{align}
    \log p(X \given \beta) =& \ \log \int \int p(X,Y, \theta \given \beta) \ \mathrm{d}\theta \ \mathrm{d}Y \nonumber \\
    =& \ \log \int \int q(Y, \theta \given \beta) \ \frac{p(X,Y,\theta \given \beta)}{q(Y, \theta \given \beta)} \ \mathrm{d}\theta \ \mathrm{d}Y \nonumber \\
    \geq& \ \int \int q(Y, \theta \given \beta) \log \frac{p(X,Y, \theta \given \beta)}{q(Y, \theta \given \beta)} \ \mathrm{d}\theta \ \mathrm{d}Y \ = \ {\cal L}(q) \, . \label{eq:elbo}
\end{align}
${\cal L}(q)$ is a function of the approximating distribution $q$ and is called the evidence lower bound. Here, the dependence on the hyperparameters is left out for notational convenience. We only maintain the dependence on $\beta$, as it is of importance in Section \ref{sec:cross}. In this framework, the objective is to find a parametric form for the variational approximation distribution $q(Y, \theta \given \beta)$ such that it matches the true distribution as well as possible \cite{bishop2006pattern}.

For computational reasons, we make the mean-field assumption that the segmentation and the likelihood parameters are independent of each other: $q(Y, \theta \given \beta) = q(Y \given \beta) \  q(\theta)$ \cite{bishop2006pattern}. The optimal form of each factor can be found by dropping terms in the lower bound that do not depend on the factor in question (as they are constants in the optimization procedure), and deriving the analytical solutions to the remaining expectations. For latent factor $q(Y \given \beta)$, terms in the numerator and denominator of (\ref{eq:elbo}) not involving $Y$ and $\beta$ are ignored, producing:
\begin{align}
    {\cal L}(q) \propto& \int \int q(Y \given \beta) \ q(\theta) \log \frac{p(X \given Y, \theta) \ p(Y \given \beta)}{q(Y \given \beta)} \ \mathrm{d}\theta \ \mathrm{d}Y  \nonumber \\ 
    =& \int q(Y \given \beta) \log \frac{\exp \big( \int q(\theta) \ \log p(X \given Y, \theta) \ + \log p(Y \given \beta) \ \mathrm{d}\theta \big)}{q(Y \given \beta)} \ \mathrm{d}Y \, . \label{eq:optfqy}
\end{align}
Note that in (\ref{eq:optfqy}), the expectation with respect to the other factor, $q(\theta)$, is moved to the numerator. It can now be seen that the latent factor is optimal when: $\log q^{*}(Y \given \beta) \propto \mathbb{E}_{\theta} \big[ \log p(X \given Y, \theta) \big] + \log p(Y \given \beta)$. Using the full Bayesian model specified in (\ref{eq:full}) and the hidden Potts from (\ref{eq:potts}), the $i$-th voxel of the segmentation factor can be written as \cite{bishop2006pattern}:
\begin{align}
    \log \ &q^{*}(y_i \given \beta) \nonumber \\
    \propto& \ \mathbb{E}_{\pi, \mu, \Lambda} \big[ \sum_{k=1}^{K} y_{ik} \log \pi_k {\cal N}(x_i \given \mu_{k}, \Lambda_{k}^{-1}) \big] + \log p(y_{i} \given y_{\delta_{i}}, \beta) \nonumber \\
    \propto& \ \sum_{k=1}^{K} y_{ik} \Big(\mathbb{E}_{\pi_k} \big[ \log \pi_k \big] + \mathbb{E}_{\mu_k, \Lambda_k} \big[ \log {\cal N}(x_i \given \mu_k, \Lambda_k^{-1}) \big] + \beta_k \sum_{j \in \delta_{ik}} y_{jk} \Big) \, , \label{eq:estep} 
\end{align} 
where
\begin{align}
    &\mathbb{E}_{\pi_k} \big[ \log \pi_k \big] = \psi(\alpha_k) - \psi \big(\sum_{k=1}^{K} \alpha_k \big) \, , \nonumber \\
    &\mathbb{E}_{\mu_k, \Lambda_k} \big[ \log {\cal N}(x_i \given \mu_k, \Lambda_k^{-1}) \big] = -\frac{D}{2} \log 2\mathrm{\pi} + \frac{1}{2}\mathbb{E}_{\Lambda_k} \big[ \log |\Lambda_{k}| \ \big] - \frac{1}{2}\mathbb{E}_{\mu_k, \Lambda_k} \big[\tilde{x}_{ik} \big] \, , \nonumber \\
    &\mathbb{E}_{\Lambda_{k}} \big[ \log | \Lambda_k | \ \big] = \ \sum_{d=1}^{D} \psi \big[(\nu_{k} + 1 - d) / 2 \big] + D \log 2 + \log | \Delta_k | \, , \nonumber \\
    &\mathbb{E}_{\mu_k, \Lambda_k} \big[\tilde{x}_{ik} \big] = \ \frac{D}{\gamma_k} + \nu_{k} (x_i - \upsilon_{k}) \Delta_{k} (x_i - \upsilon_{k})^{\top} \nonumber \, ,
\end{align}
and $\tilde{x}_{ik} = (x_i - \mu_{k}) \Lambda_{k} (x_i - \mu_k)^{\top}$. $\psi$ refers to the digamma function. 

In Equation \ref{eq:estep}, we recognize a multinomial distribution: $\log q^{*}(y_i \given \beta) \propto \ \sum_{k}^{K} y_{ik} \log r_{ik}$. The proportionality is due to the ignored terms. As these terms only serve to normalize the probabilities to the $[0,1]$ interval, we can replace their computation by the following normalization: $\rho_{i} = r_{ik} / \sum_{k=1}^{K} r_{ik} $ \cite{bishop2006pattern}. The $\rho_{ik}$ are called the responsibilities, referring to the probability for the $i$-th voxel to belong to the $k$-th class. Note that $\beta$ has not been integrated out. It will be estimated in a cross-medical center fashion (see Section \ref{sec:cross}).

Similar steps are taken to compute an optimal form for $q(\theta)$. This time, we ignore all terms in the ratio in (\ref{eq:elbo}) that do not depend on $\theta$:
\begin{align}
    {\cal L}(q) \propto& \int \int q(Y \given \beta) \ q(\theta) \log \frac{p(X \given Y, \theta) \ p(\theta)}{q(\theta)} \ \mathrm{d}\theta \ \mathrm{d}Y \nonumber \\ 
    =& \int q(\theta) \log \frac{\exp \big( \int q(Y \given \beta) \ \log p(X \given Y, \theta) \ + \log p(\theta) \ \mathrm{d}Y \big)}{q(\theta)} \ \mathrm{d}\theta \, . \label{eq:optfqz}
\end{align}
The factor $q(\theta)$ is optimal when: $\log q^{*}(\theta) \propto \mathbb{E}_{Y} \big[ \log p(X \given Y, \theta) \ \big] + \log p(\theta)$. This is a well-known result (the choice of a Gaussian likelihood with conjugate priors is made often) and extensive derivations are widely available \cite{bishop2006pattern}. It produces the following update equations:
\begin{align}
    \alpha_{k} =& \ \alpha_{0k} + S^{0}_k \ , \quad \quad 
    \gamma_{k} = \ \gamma_{0k} + S^{0}_k \ , \quad \quad 
    \nu_{k} = \ \nu_{0k} + S^{0}_k \ \, , \nonumber \\
    \upsilon_k =& \ (\gamma_{0k} \upsilon_{0k} + S^{1}_k) \ / \ (\gamma_{0k} + S^{0}_k) \ \, , \nonumber \\
    \Delta_k^{-1} =& \ \Delta^{-1}_{0k} + S^{2}_{k} + \frac{\gamma_{0k}S^{0}_{k}}{\gamma_{0k} + S^{0}_{k}} (S^{1}_{k} - \upsilon_{k})(S^{1}_{k} - \upsilon_k)^{\top} \ \, , \label{eq:mstep}
\end{align}
where parameters with the subscript $0$ belong to the priors and
\begin{align}
    S^0_k = \sum_{i=1}^{N} \rho_{ik} \ , \quad \quad
    S^1_k = \sum_{i=1}^{N} \rho_{ik} x_i \ , \quad \quad 
    S^2_k = \sum_{i=1}^{N} \rho_{ik} (x_i - S^1_k)(x_i - S^1_k)^{\top} \ . \nonumber
\end{align}
Note that the smoothness parameters $\beta$ affect these hyperparameter estimates through the estimates of the responsibilities $\rho_{ik}$. 

Inference consists of iteratively computing the responsibilities based on the current posterior hyperparameters followed by updating the posterior hyperparameters given the new responsibilities. This procedure, known as variational Bayes, is halted when the change in values between iterations is smaller than a set threshold \cite{bishop2006pattern,mcgrory2009variational,blaiotta2016variational}.

\subsection{Semi-supervised model} \label{sec:semi}
Unsupervised models are limited by the fact that cluster assignments are not tied to tissue labels. Manually labeling one voxel per tissue overcomes this limitation. In order to incorporate the given voxel labels, a split in the likelihood function between labeled samples and unlabeled samples needs to be introduced \cite{krijthe2014implicitly}:
\begin{align}
    p(X, \tilde{Y} \given Y; \ \theta) = \prod_{j \in {O}} \prod_{i \centernot\in {O}} \prod_{k=1}^{K} \big[ \pi_k \mathcal{N}(x_j \given \mu_{k}, \Lambda_{k}^{-1})\big]^{\tilde{y}_{jk}} \big[ \pi_{k} & \mathcal{N}(x_i \given \mu_{k}, \Lambda_{k}^{-1})\big]^{y_{ik}} \, , \nonumber
\end{align}
where $\tilde{Y}$ are the observed labels, $Y$ are the unobserved labels and $O \subset [1, \dots N]$ is the subset of indices that are observed.

To derive a semi-supervised hidden Potts Gaussian mixture requires substituting the unsupervised likelihood from (\ref{eq:likelihood}) in the Bayesian model in (\ref{eq:full}) with the above semi-supervised likelihood. Using the same derivations as throughout Section \ref{sec:method}, this results in equivalent update equations for both the $q(Y \given \beta)$ and $q(\theta)$ with the following exception: the responsibilities of the observed voxels are fixed to $\rho_{ik} = 1.0$ if class $k$ was observed, and to $\rho_{im} = 0.0$ for $m \neq k$. These responsibilities remain fixed throughout the variational optimization procedure.

\subsection{Initialization of posterior hyperparameters}
Variational inference is a form of non-convex optimization, which means that different initializations lead to different local optima. Several initializations for variational mixture models have been proposed, most notably $k$-means for unsupervised Gaussian mixtures \cite{nasios2006variational}. In that case, the responsibilities $\rho$ of a point are set by the negative exponential of the distance to each cluster center.

For the semi-supervised model, the responsibilities can be initialized based on the distance to the given labeled voxels. This corresponds to $k$-nearest-neighbour classifier. As long as the number of labeled pixels is small, this remains computationally efficient.

\section{Cross-center empirical Bayes} \label{sec:cross}
The Potts model can be fit to other segmentations using a maximum likelihood approach. First, it is treated as a likelihood function in its own right, with $Y$ as the observed variable dependent on the smoothing parameters $\beta$. Using the log-likelihood, the estimator becomes:
\begin{align}
    \hat{\beta} = \ \underset{\beta \in \mathbb{R}^{+}}{\arg \max} \ \sum_{i=1}^{N} \log p(y_{i} \given y_{\delta_{i}}, \beta) \, . \nonumber
\end{align}
This log-likelihood function is convex in $\beta$, which means the optimal smoothing parameters can be obtained using gradient descent. Its partial derivative with respect to $\beta$ is:
\begin{align}
    \frac{\partial}{\partial \beta} \log p(y_{i} \given y_{\delta_{i}}, \beta) 
    =& \ \sum_{k=1}^{K} y_{ik} \bar{y}_{ik} - \sum_{l=1}^{K} \bar{y}_{il} \exp \big(\beta_{l} \bar{y}_{il} \big) / \Big[ \sum_{m=1}^{K} \exp \big(\beta_{m} \bar{y}_{im} \big) \Big] \, . \nonumber
\end{align}
where $\bar{y}_{ik} = \sum_{j \in \delta_{i}} y_{jk}$. Using a constrained optimization procedure, where all $\beta$'s are lower bounded by $0$, we can obtain a point estimate for each $\beta_k$.

\section{Experiments} \label{sec:experiments}
We perform a series of cross-center brain tissue segmentation experiments. The goal is to assign each pixel in the MR image the label "background", "cerebro-spinal fluid", "gray matter", or "white matter". One data set will act as the source and another as the target. For the sake of comparison, we include single-center experiments, where $\beta$ cannot be learned and is subsequently set to $0.1$. All classification errors are computed using the brain mask. In other words, we ignore all mistakes in the skull and outlying regions. With the unsupervised models, each cluster is interpreted as one of the tissues, so that classification errors can be computed. To test the performance gain that can be achieved with a small amount of supervision, we provide the semi-supervised models with 1 voxel label per tissue from the target image, sampled at random. The experiments are repeated 10 times.

\subsection{Data sets}
\begin{figure}
    \centering
    \includegraphics[height=100px]{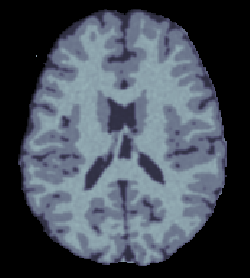} \
    \includegraphics[height=100px]{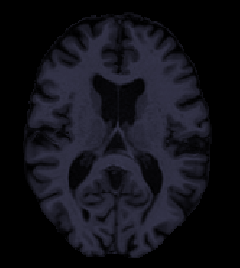} \
    \includegraphics[height=100px]{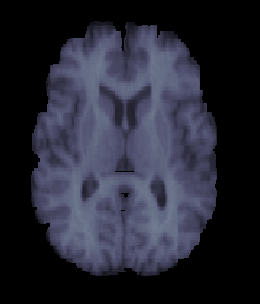}
    \caption{Example scans. (Left) Brainweb1.5T, (middle) MRBrainS13 and (right) IBSR.}
    \label{fig:examples}
\end{figure}

We will make use of 3 publicly available data sets: Brainweb1.5T, MRBrainS13, and IBSR. Each data set originates from one medical center. Brainweb1.5T is based on 20 realistic phantoms from Brainweb \cite{aubert2006twenty} and an MRI simulator (SIMRI; \cite{benoit2005simri}). The simulator was set to use TE, TR and flip angle parameters of the 1.5T scanner in the Rotterdam Scan Study \cite{ikram2015rotterdam}. MRBrainS13 is a grand challenge for medical image tissue segmentation methods containing 5 scans for training \cite{mendrik2015mrbrains}. The scans are 3T and have been fully manually annotated. IBSR is a classical data set of 18 patients and is automatically segmented but manually corrected \cite{rohlfing2012image}. Skulls are stripped off in all scans. Figure \ref{fig:examples} visualizes examples from these sets.

\subsection{Segmentation methods}
We will compare the following methods: firstly, a U-net consisting of a mirrored VGG16 architecture pre-trained on ImageNet and fine-tuned on labeled data from the source medical center \cite{ronneberger2015u,simonyan2014very}. This method represents the performance of a state-of-the-art tissue segmentation model without taking center-based variation into account. Secondly, we take both an unsupervised and a semi-supervised variational Gaussian mixture model (UGM, SGM), initialized using $k$-means and $1$-nearest-neighbours respectively. Thirdly, an unsupervised and a semi-supervised hidden Potts Gaussian mixture (UHP, SHP) are taken, also initialized using $k$-means and $1$-nearest-neighbours. Comparing these with the previous two models shows the influence of smoothing the segmentations. Lastly, we train a $1$-nearest-neighbours (1NN) based on the labeled voxels (1 per tissue) in the target image, as a baseline supervised tissue classifier. The maximum number of training iterations is set to 30 for all methods.

\subsection{Results}
We present mean classification errors (with standard errors of the means over 10 repetitions) of each method in Table \ref{tab:pairs}. Firstly, comparing the performances of UHP and SHP in the experiments off the diagonal (multi-center) with their performances on the diagonal (single-center) shows that the learned smoothness parameters are more effective than the chosen ones. Secondly, the errors of the hidden Potts models versus the standard Gaussian mixtures tend to be lower or similar (UHP $<=$ UGM and SHP $<=$ SGM). Thirdly, the semi-supervised models tend to outperform the unsupervised ones (SGM $<$ UGM and SHP $<$ UHP). Taken the performance of 1NN into account, it shows that even 1 label per tissue is very informative. U-net performs poorly as it is not aware of the intensity and contrast shifts between data sets.
\begin{table}
\setlength{\tabcolsep}{10pt} 
\renewcommand{\arraystretch}{1.0} 
\caption{Mean classification error and standard errors of the means (in brackets) of each of the segmentation models on all pairwise combinations of one data set as the source (rows) and another as the target (columns).}
\label{tab:pairs}
    \begin{tabular}{l | c | c | c | c }
                                    & Methods & Brainweb1.5T & MRBrainS13 & IBSR  \\
    \midrule
    \multirow{6}{6em}{Brainweb1.5T} & U-net  & -            & 0.448 (.008) & 0.384 (.019) \\
                                    & 1NN   & 0.117 (.027) & 0.288 (.019) & 0.518 (.075) \\
                                    & UGM   & 0.142 (.044) & 0.339 (.023) & 0.525 (.088) \\
                                    & SGM   & 0.116 (.031) & 0.268 (.017) & 0.527 (.089) \\
                                    & UHP   & 0.147 (.053) & 0.337 (.023) & 0.511 (.087) \\
                                    & SHP   & 0.117 (.032) & 0.253 (.017) & 0.519 (.078) \\
    \midrule
    \multirow{6}{6em}{MRBrainS13} & U-net    & 0.257 (.003) & -            & 0.589 (.022) \\ 
                                  & 1NN     & 0.103 (.010) & 0.282 (.018) & 0.513 (.075) \\
                                  & UGM     & 0.112 (.024) & 0.345 (.022) & 0.521 (.090) \\
                                  & SGM     & 0.102 (.011) & 0.282 (.021) & 0.502 (.093) \\
                                  & UHP     & 0.119 (.030) & 0.344 (.018) & 0.507 (.091) \\
                                  & SHP     & 0.102 (.008) & 0.277 (.020) & 0.503 (.076) \\
    \midrule
    \multirow{6}{6em}{IBSR}   & U-net & 0.334 (.007) & 0.425 (.015) & - \\ 
                              & 1NN  & 0.102 (.005) & 0.282 (.064) & 0.492 (.034) \\
                              & UGM  & 0.125 (.037) & 0.369 (.071) & 0.508 (.039) \\
                              & SGM  & 0.103 (.015) & 0.260 (.043) & 0.502 (.039) \\
                              & UHP  & 0.123 (.023) & 0.350 (.068) & 0.509 (.041) \\
                              & SHP  & 0.103 (.008) & 0.255 (.049) & 0.496 (.041) 
    \end{tabular}
\end{table}

\begin{figure}
    \centering
    \includegraphics[width=.24\textwidth]{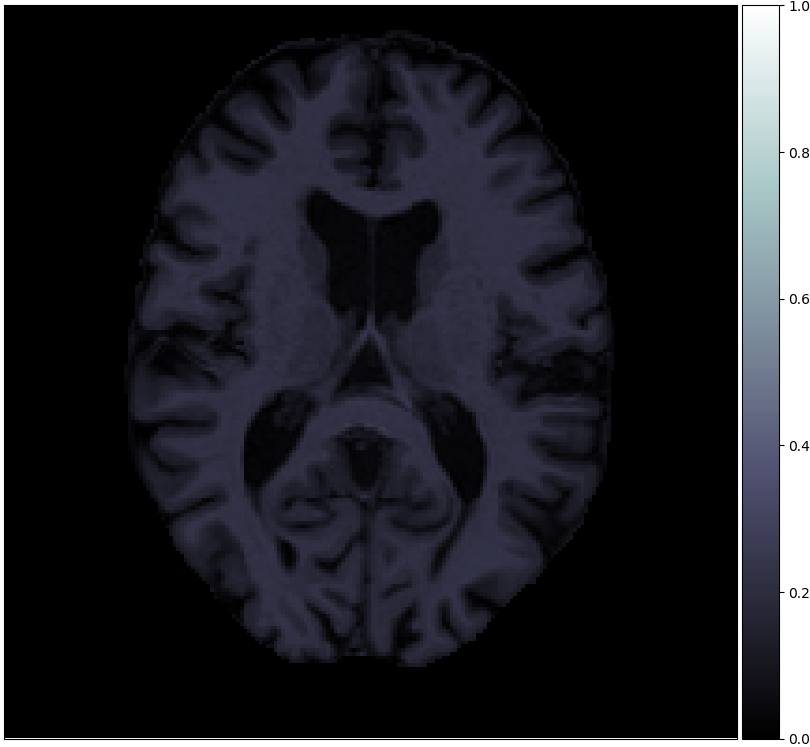}
    \includegraphics[width=.24\textwidth]{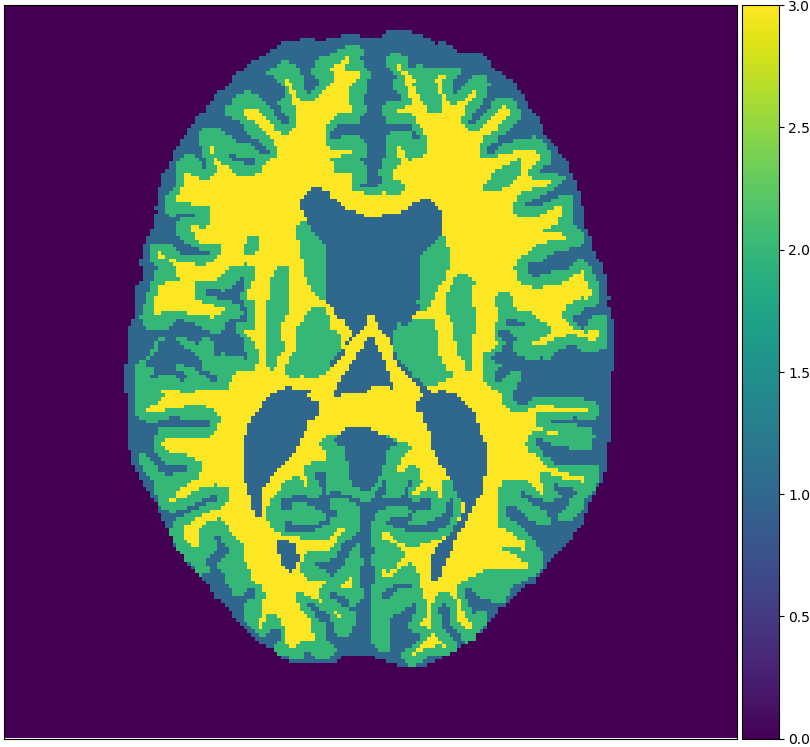}
    \includegraphics[width=.24\textwidth]{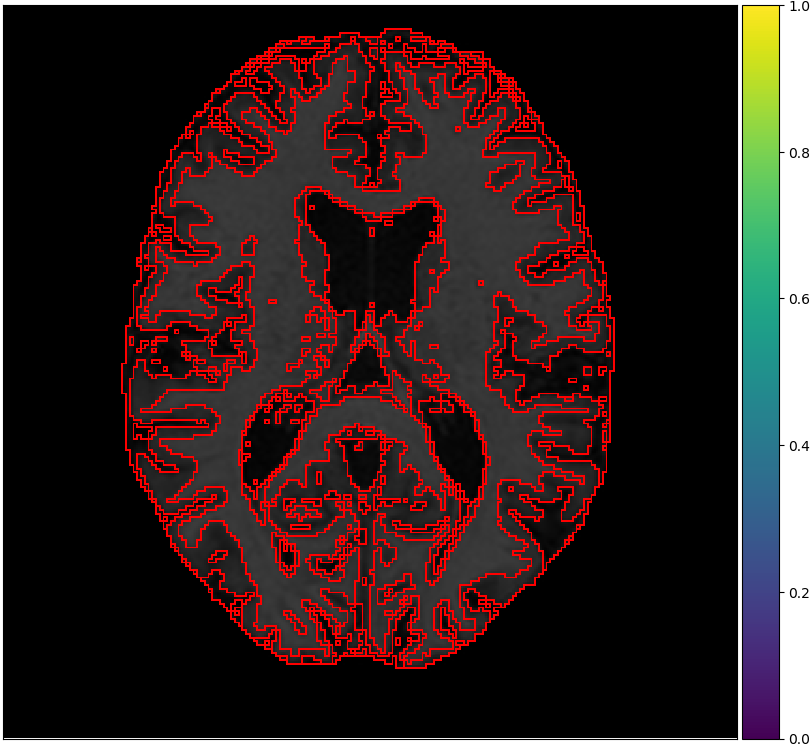}
    \includegraphics[width=.24\textwidth]{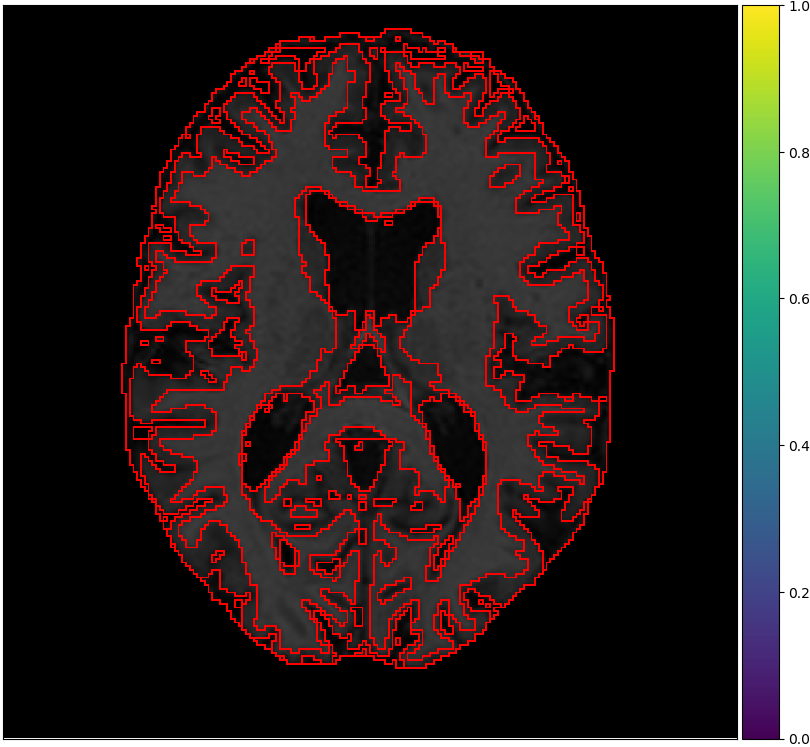}\\
    \includegraphics[width=.24\textwidth]{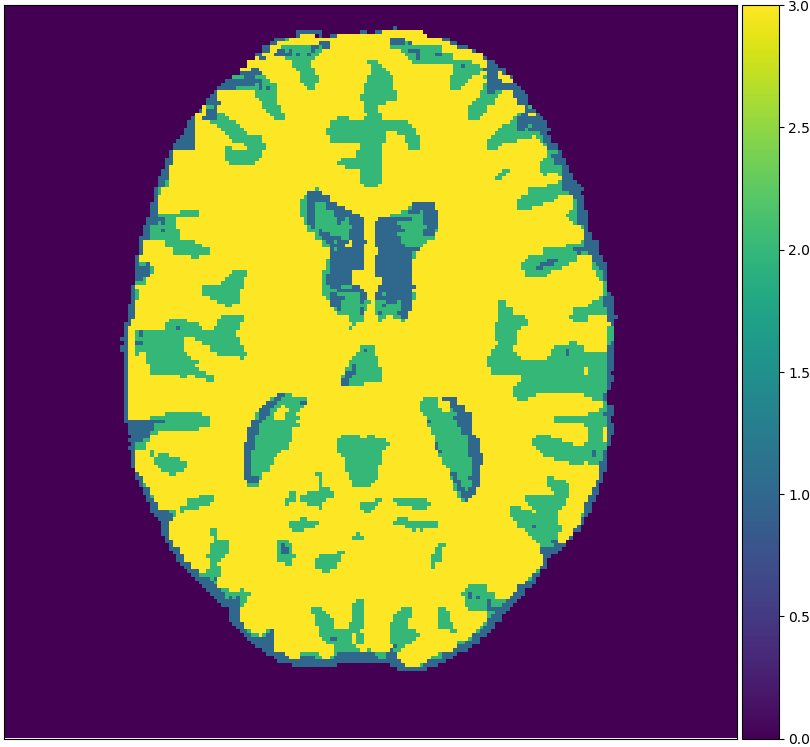}
    \includegraphics[width=.24\textwidth]{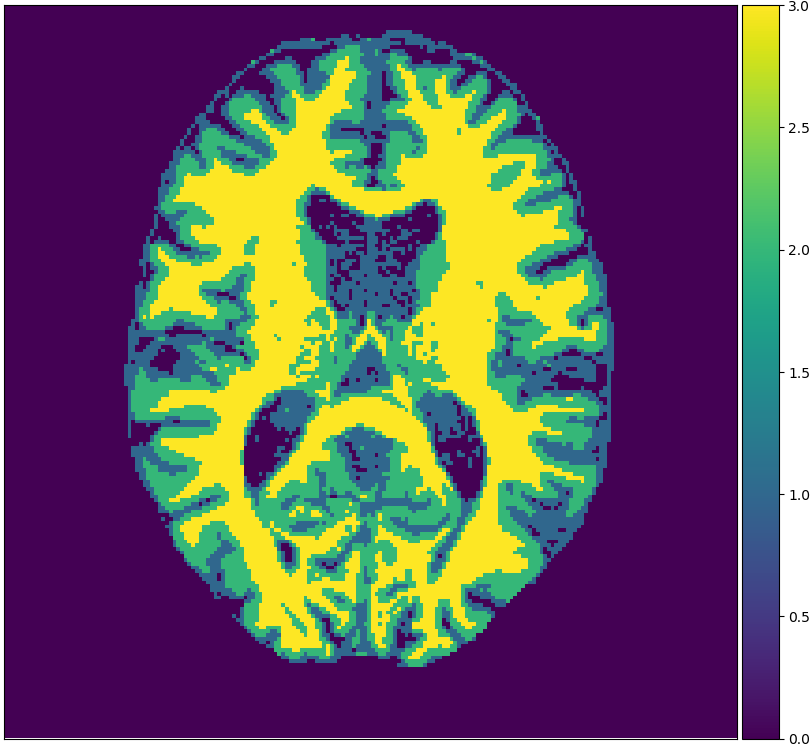}
    \includegraphics[width=.24\textwidth]{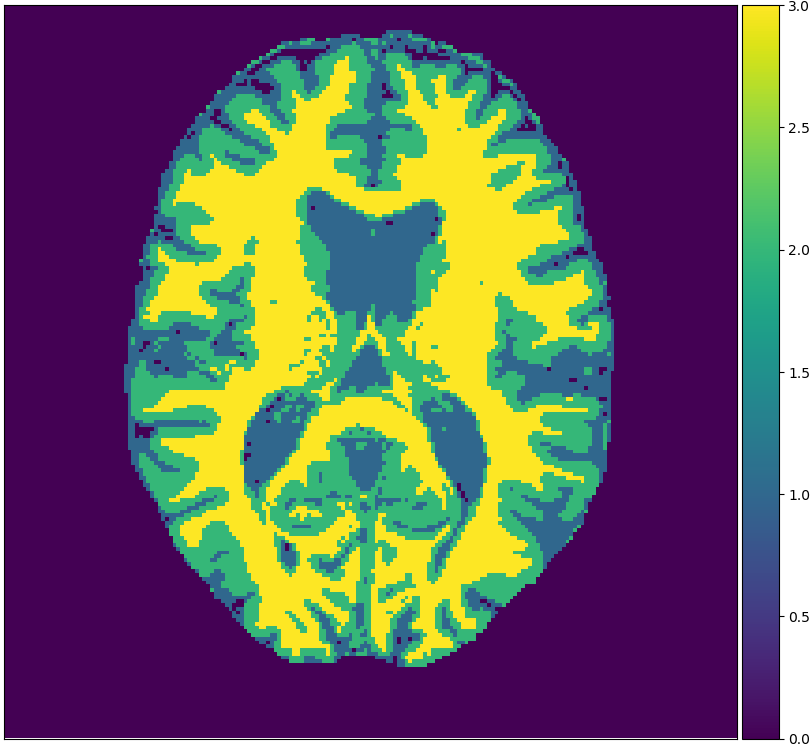}
    \includegraphics[width=.24\textwidth]{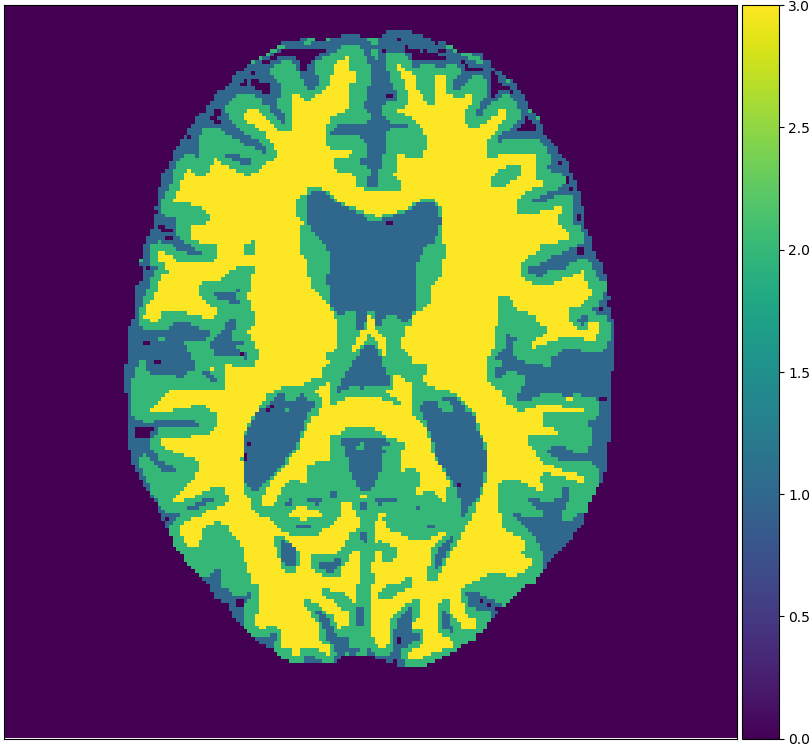}
    \caption{Segmentations with Brainweb1.5T as source and MRBrainS13 as target data. Top row, from left to right: original scan, true segmentation, unsupervised Gaussian mixture (UGM), unsupervised hidden Potts (UHP). Bottom row: U-net, $1$-nearest-neighbour, semi-supervised Gaussian mixture (SGM), semi-supervised hidden Potts (SHP). Purple = background, blue = cerebro-spinal fluid, green = gray matter and yellow = white matter.}
    \label{fig:examples_mrb}
\end{figure}

Figure \ref{fig:examples_mrb} shows examples of each segmentation method on the MRBrainS13 data set, with Brainweb1.5T as the source center. For the unsupervised models we only show boundaries between clusters, to indicate that interpretation remains a necessary step. A couple of observations can be made: firstly, the hidden Potts models produce smoother segmentations. Secondly, the U-net over-predicts white matter in the whole image. Thirdly, the 1-nearest-neighbours classifier over-predicts background voxels in fluid regions. Lastly, all methods favour white matter over gray matter in ambiguous regions.

\section{Discussion} \label{sec:discuss}
Although segmentations remain relatively consistent across medical centers compared to the scans, annotator variation can be quite large. This is especially true if medical centers teach different annotation protocols. To account for this type of variation, it would be more appropriate to capture the uncertainty in smoothness and infer the posterior over $\beta$ \cite{mcgrory2009variational}.

In our formulation, the hidden Potts-MRF acts as a spatial regularizer on the responsibilities estimated by the variational mixture model. Spatial regularizers are not uncommon, but are often employed on the observed data: most models incorporate information on the smoothness in $X$ to estimate $Y$. Here, we explicitly look at smoothness in $Y$.

A limitation of the current model is that it is not appropriate for abnormality or pathology detection. That would require a different number of components for images \emph{with} pathologies versus images \emph{without} pathologies. However, it should be possible to extend variational Gaussian mixture models to incorporate a variable amount of components. In that case, the component weights are not modeled using a Dirichlet distribution, but a Dirichlet process \cite{blei2006variational}. 

\section{Conclusion} \label{sec:conclusion}
We proposed to tackle center-specific variation in medical imaging data sets with Bayesian transfer learning. We fitted a spatial smoothness prior on the segmentations produced in one medical center and used this informative prior to perform brain tissue segmentation at the target center. Our results show improvements over non-spatially smoothed segmentations, and improvements with learned smoothness parameters over chosen ones.
\bibliographystyle{splncs04}
\bibliography{kouw_ipmi19a}
\end{document}